\UseRawInputEncoding  

\documentclass[twocolumn,final]{elsarticle}

\usepackage{lineno}
\usepackage{framed,multirow}
\usepackage{amssymb}
\usepackage{latexsym}

\usepackage{url}
\usepackage{xcolor}

\usepackage{graphicx}
\usepackage{amsmath}

\usepackage{verbatim}
\usepackage{booktabs}
\usepackage{multirow}
\usepackage{bm}
\usepackage{array}
\usepackage{graphicx}
\usepackage{epstopdf}
\usepackage{amssymb}

\usepackage{makecell}
\usepackage{tabularx} 
\usepackage{booktabs}   
\usepackage{mathrsfs} 
\usepackage{multirow} 
\usepackage{amsfonts} 
\usepackage{wrapfig}  
\usepackage{cite}
\usepackage{footnote}
\usepackage{xcolor}
\usepackage{caption}
\usepackage{subcaption}

\usepackage{bbding}
\usepackage{times}
\usepackage{latexsym} 
\usepackage{algorithmic}
\usepackage{algorithm}
\usepackage{dsfont}

\usepackage{textcomp}
\usepackage{array}
\usepackage{amsmath,amsfonts}

\usepackage{dirtytalk}
\usepackage{latexsym}

\usepackage{algorithm,algorithmic}

\usepackage{url}
\usepackage{xcolor}
\usepackage[hidelinks]{hyperref}

\def\x{{\bf x}}

\def\0{{\bf 0}}
\def\1{{\bf 1}}

\definecolor{myblue}{RGB}{0,143,156}
\newcommand{\citeblue}[1]{\textcolor{myblue}{\citep{#1}}}
\newcommand{\citetblue}[1]{\textcolor{myblue}{\citet{#1}}}

\newcommand{\myref}[1]{\textcolor{myblue}{\ref{#1}}}

\biboptions{authoryear}

\usepackage{geometry}
\begin{document}
	
	\begin{frontmatter}
		\title{SPLAL: Similarity-based pseudo-labeling with alignment loss for semi-supervised medical image classification}

\author[mymainaddress]{Md Junaid Mahmood}\ead{md\_j@ec.iitr.ac.in}
\author[mymainaddress]{Pranaw Raj}\ead{pranaw\_r@cs.iitr.ac.in}
\author[mymainaddress]{Divyansh Agarwal}\ead{divyansh\_a@ee.iitr.ac.in}
\author[mymainaddress]{Suruchi Kumari}\ead{suruchi\_k@cs.iitr.ac.in}
\author[mymainaddress]{Pravendra~Singh \corref{mycorrespondingauthor}}
\ead{pravendra.singh@cs.iitr.ac.in}
\cortext[mycorrespondingauthor]{Corresponding author: Pravendra Singh}

\address[mymainaddress]{Department of Computer Science and Engineering, Indian Institute of Technology Roorkee, India}

\begin{abstract}
Medical image classification is a challenging task due to the scarcity of labeled samples and class imbalance caused by the high variance in disease prevalence. Semi-supervised learning (SSL) methods can mitigate these challenges by leveraging both labeled and unlabeled data. However, SSL methods for medical image classification need to address two key challenges: (1) estimating reliable pseudo-labels for the images in the unlabeled dataset and (2) reducing biases caused by class imbalance.
In this paper, we propose a novel SSL approach, SPLAL, that effectively addresses these challenges. SPLAL leverages class prototypes and a weighted combination of classifiers to predict reliable pseudo-labels over a subset of unlabeled images. Additionally, we introduce alignment loss to mitigate model biases toward majority classes. To evaluate the performance of our proposed approach, we conduct experiments on two publicly available medical image classification  benchmark datasets: the skin lesion classification (ISIC 2018) and the blood cell classification dataset (BCCD). The experimental results empirically demonstrate that our approach outperforms several state-of-the-art SSL methods over various evaluation metrics. Specifically, our proposed approach achieves a significant improvement over the state-of-the-art approach on the ISIC 2018 dataset in both Accuracy and F1 score, with relative margins of 2.24\% and 11.40\%, respectively. Finally, we conduct extensive ablation experiments to examine the contribution of different components of our approach, validating its effectiveness. 
\end{abstract}

\end{frontmatter}

\section{Introduction}
\label{sec:introduction}

The field of computer-aided diagnosis plays a vital role in enhancing diagnostic efficiency and reducing the likelihood of incorrect diagnosis, making it an area of considerable importance and interest within the research community. In recent times, various research works on deep learning have shown outstanding results in medical image classification \citeblue{zhang2019attention, huang2019blood, sun2021skin}. However, the performance of such approaches are dependent upon the existence of large labeled datasets \citeblue{he2016deep}. In a real-world scenario, labeling of medical images for training deep learning models is an expensive option. This is because the labeling of high-quality data is time-consuming and requires a high level of proficiency from medical experts \citeblue{litjens2017survey}.

Consequently, we have a relatively lower availability of quality labeled dataset for most diseases. However, there is
always a scope of exploring unlabeled images from clinics and
hospitals databases. Semi-Supervised Learning (SSL) \citeblue{rosenberg2005semi, grandvalet2004semi, berthelot2019mixmatch, sohn2020fixmatch} offers a
means to utilize unlabeled data for training, thus minimizing the
need for a large labeled dataset. 

Pseudo-labeling is a technique in semi-supervised learning to generate pseudo-labels using the predictions on unlabeled data, which are then utilized during the training process. \citetblue{lee2013pseudo} first proposed a pseudo-labeling approach in which a model trained only over labeled images is used to predict pseudo-labels for the unlabeled images. In such a scenario, the class with the highest prediction probability is chosen as the pseudo-label for the corresponding unlabeled image. However, pseudo-labeling approaches that rely solely on the model's output can cause confirmation bias \citeblue{arazo2020pseudo}. Moreover, training the model using incorrect pseudo-labels can lead to increased confidence in incorrect predictions, resulting in a decrease in the model's classification performance on unseen samples.

\begin{figure}
     \centering
     \begin{subfigure}[b]{0.3\textwidth}
         \centering
         \includegraphics[width=\textwidth]{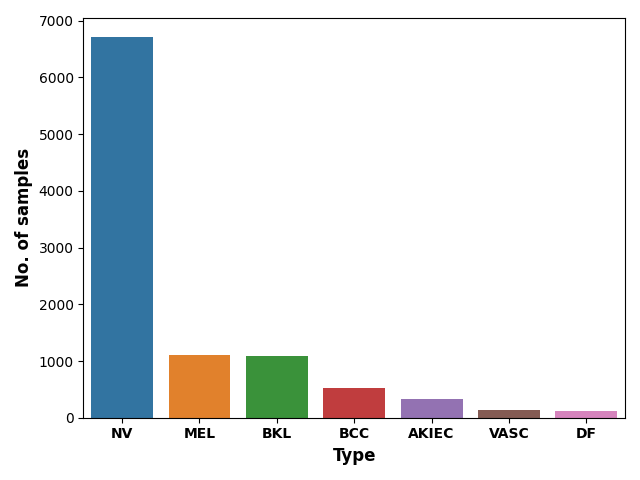}
         \caption{Images per label in ISIC 2018 dataset.}
         \label{fig:isic18_distribution}
     \end{subfigure}
     \hfill
     \begin{subfigure}[b]{0.3\textwidth}
         \centering
         \includegraphics[width=\textwidth]{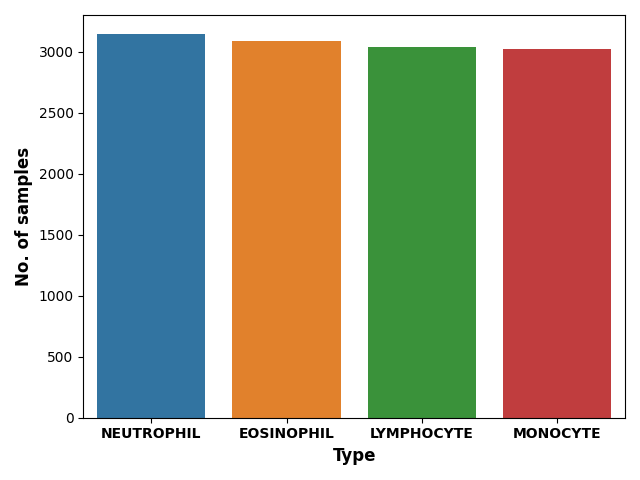}
         \caption{Images per label in BCCD.}
         \label{fig:bccd_distribution}
     \end{subfigure}
     \hfill
        \caption{Variance in data distribution in (a) Skin lesion classification dataset (ISIC 2018) \citeblue{tschandl2018ham10000, codella2019skin} (highly imbalanced) and (b) Blood cell classification dataset (BCCD)  \citeblue{bloodCellImages} (relatively balanced).}
        \label{fig:both_figure}
\end{figure}
\begin{figure*}[t!]
     \centering
     \begin{subfigure}[b]{0.97\textwidth}
         \centering
         \includegraphics[width=\textwidth]{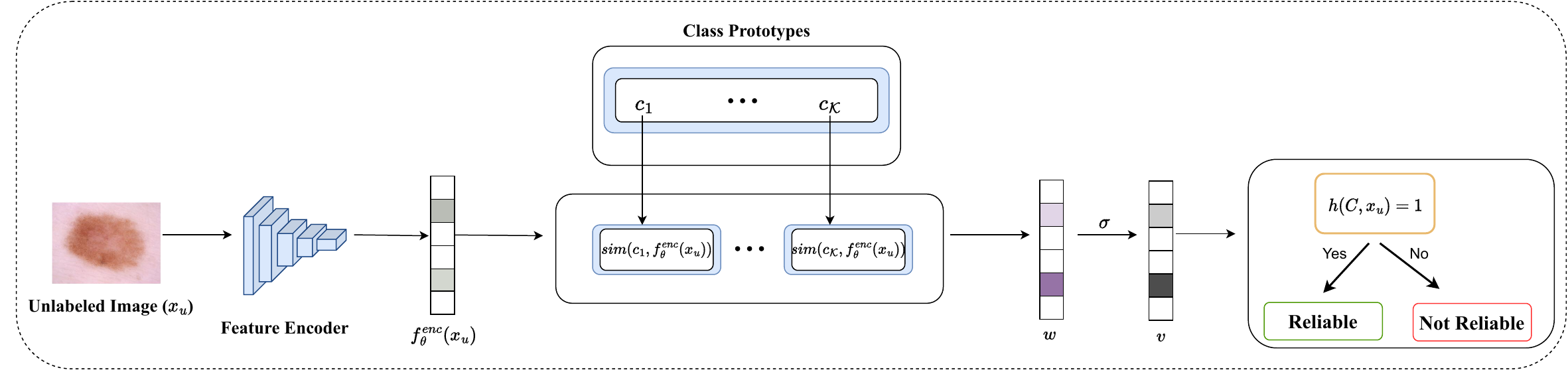}
         \caption{Reliable sample selection from unlabeled dataset using similarity with class prototypes as the criterion.}
         \label{fig:reliable unlabeled sample}
     \end{subfigure}
     \hfill
     \vspace{0.3cm}
     \begin{subfigure}[b]{0.95\textwidth}
         \centering
         \includegraphics[width=\textwidth]{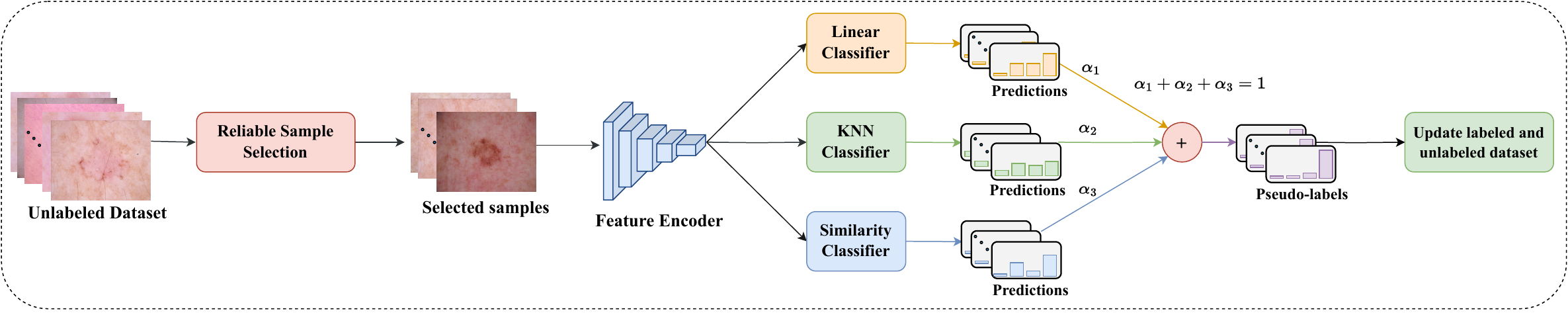}
         \caption{SPLAL pseudo-labeling framework using a weighted combination of a similarity classifier, a KNN classifier and a linear classifier.}
         \label{fig:pseudo-label calc}
     \end{subfigure}
     \hfill
     \vspace{0.3cm}
        
        \begin{subfigure}[b]{0.8\textwidth}
         \centering
         \includegraphics[width=\textwidth]{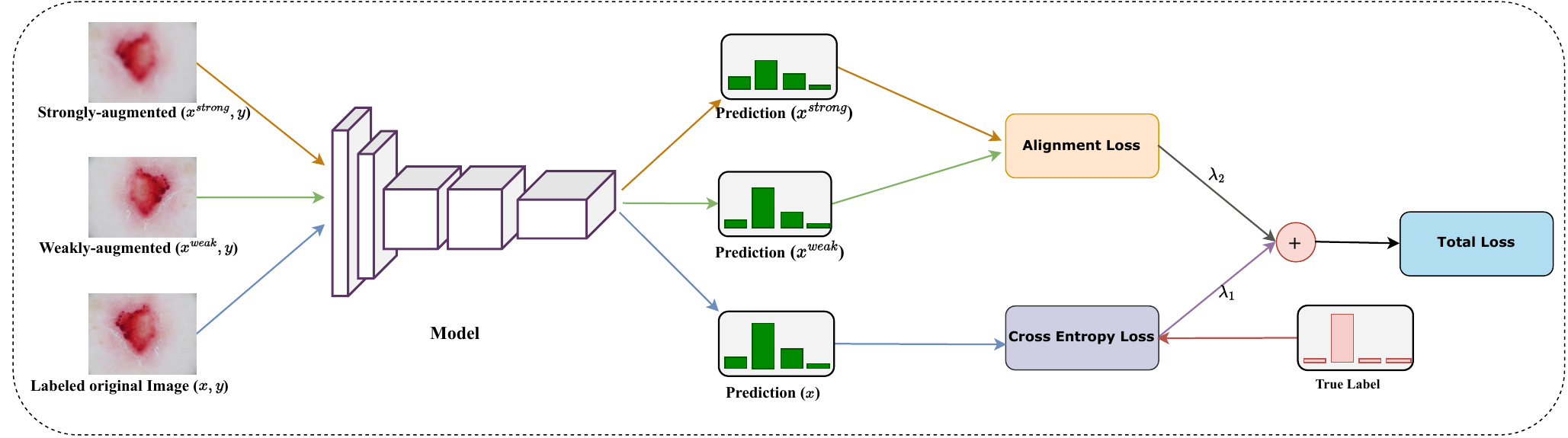}
         \caption{SPLAL optimization using weighted sum of classification loss and an alignment loss.}
         \label{fig:splal optim}
     \end{subfigure}
     \hfill
        \caption{Similarity-based Pseudo-Labeling with Alignment Loss (SPLAL) approach. The approach is divided into the following iterative steps: 1.) Reliable sample selection- In (a), we depict our novel method for estimating the reliability of an unlabeled sample for pseudo-labeling. For every unlabeled sample ($x_u$), we calculate the similarity (using $sim(.)$) of its feature vector ($f^{\text{enc}}_\theta(x_u)$) with the class prototype ($c_k$) of every class and store it in vector $w$. Then, the softmax function ($\sigma$) is applied on $w$ to obtain vector $v$. If the unlabeled image meets the criterion $h(.)$ (explained in Sec.\myref{sec:prototype}), we consider that unlabeled sample to be reliable and estimate its pseudo-label. 2.) SPLAL pseudo-labeling framework- In (b), we depict our novel method for estimating the pseudo-label of a reliable unlabeled sample. For every reliable unlabeled image $x_u$, we take the weighted sum of the prediction from a similarity classifier, a KNN classifier, and a linear classifier to estimate its pseudo-label (described in Sec.~\myref{sec:pseudo_labeling}). 3.) SPLAL optimization- In (c), we depict the total loss in SPLAL optimization. For a training image $x$, its two augmentations -- a weak ($x^{weak}$) and a strong ($x^{strong}$) are generated and passed through the model, and their corresponding predictions are obtained. Alignment loss is calculated between the two predictions. Classification loss (\textit{cross entropy}) is calculated between the model's prediction for the original training sample $x$ and its label (or pseudo-label) $y$. The total loss is defined as the weighted sum of these two losses (given in Eq.~\eqref{eq:eqL1}).}
    \label{fig:loss}
\end{figure*}

In this paper, we propose \textbf{Similarity-based Pseudo-Labeling with Alignment Loss (SPLAL)} -- a novel SSL approach that makes better use of the information available from unlabeled data to improve the classification performance of a deep learning model. Our approach maintains a prototype of every class generated using a fraction of the most recently viewed training samples of a class. This prototype generation method is inspired by DASO \citeblue{oh2022daso}. To select reliable unlabeled samples, SPLAL uses the similarity of these samples with the class prototypes. We predict the pseudo-label for the selected reliable unlabeled samples using a similarity classifier, a KNN classifier, and a linear classifier. Our reliable sample selection method (depicted by Fig.~\myref{fig:reliable unlabeled sample}) and pseudo-labeling approach (depicted by Fig.~\myref{fig:pseudo-label calc}) are described in detail in Sec.~\myref{sec:prototype} and Sec.~\myref{sec:pseudo_labeling} respectively. The selection of reliable samples using similarity with class prototypes as a criterion and its pseudo-labeling using a weighted combination of classifiers ensures that our model learns to classify various subtle representations of samples for every class correctly. The improvement in performance due to our novel reliable sample selection method and pseudo-labeling approach is empirically justified in Sec.~\myref{sec:informationThreshold} and Sec.~\myref{sec:ensemle_pl_prediction_effect}, respectively.

Furthermore, as depicted by Fig.~\myref{fig:isic18_distribution}, medical image datasets commonly have an imbalanced data distribution. Despite using a reliable pseudo-labeling method, the imbalanced class distribution can bias the model’s predictions toward the majority classes. Thus, a mechanism is required to ensure that the model's prediction towards all the classes is consistent, especially the minority classes. To ensure this, our SPLAL method uses an alignment loss which utilizes weak and strong augmentation of the input image and is directly proportional to the difference between the model's prediction for the two augmentations. The improvement in performance due to alignment loss incorporation is empirically justified in Sec.~\myref{sec:regulatization_effect}. 

We evaluate SPLAL on two publicly available medical image classification benchmark datasets, namely the skin lesion classification (ISIC 2018) dataset \citeblue{tschandl2018ham10000, codella2019skin} and the blood cell classification dataset (BCCD) \citeblue{bloodCellImages}. Our method outperforms several state-of-the-art SSL methods over various evaluation metrics as shown in Table~\myref{table:isic18_main} and Table~\myref{table:bccd_main}. Furthermore, we conducted extensive ablation experiments on the ISIC 2018 dataset to understand the contribution of each component of our approach, which is discussed in detail in Sec.~\myref{sec:ablation_study}.

Our approach is illustrated in Fig.~\myref{fig:loss}. The detailed explanation of our approach is given in Sec.~\myref{sec:methods}. Our contributions can be summarized as follows:
\begin{itemize}
    \item We propose a novel approach for reliable sample selection from unlabeled dataset using class prototypes.
    \item We propose a novel method for predicting pseudo-labels from unlabeled samples using a weighted combination of a similarity classifier, a KNN classifier, and a linear classifier, which generates high-quality pseudo-labels and improves the accuracy of SSL.
    \item Incorporating an alignment loss using weak and strong augmentations of an image to enforce consistent predictions and empirically demonstrate that this loss mitigates model biases toward majority classes. 
    \item We perform experiments on multiple benchmark datasets to show that our approach significantly outperforms several state-of-the-art SSL methods over various evaluation metrics. We also perform extensive ablation experiments to validate the different components of our approach.
\end{itemize}

\section{Related Work}
\label{sec:related_work}

\subsection{Semi-supervised learning}
Semi-supervised learning (SSL) aims to leverage both labeled and unlabeled data to improve the performance of deep learning models \citeblue{bai2017semi}. One of the popular techniques for utilizing unlabeled data is pseudo-labeling, which involves generating labels for unlabeled samples using a model's predictions. Typically, a confidence threshold is set to select only high-confidence predictions for use as pseudo-labels, which can then be used as training samples to train the model. \citetblue{lee2013pseudo} first proposed a pseudo-labeling method, in which a neural network model trained solely on labeled samples is used to generate pseudo-labels. However, pseudo-labeling solely based on model outputs can result in confirmation bias \citeblue{arazo2020pseudo}.

In recent years, there has been significant research on pseudo-labeling from two main perspectives. Firstly, several works \citeblue{li2021comatch, hu2021simple, tarvainen2017mean, saito2021openmatch, berthelot2019mixmatch} have proposed methods to enhance the consistency of predictions made on samples from different viewpoints. The Mean-teacher approach \citeblue{tarvainen2017mean} enforces similarity between predictions of the student model and its momentum teacher model, while MixMatch \citeblue{berthelot2019mixmatch} suggests a technique to reduce the discrepancy among multiple samples that are augmented using mixup. \citetblue{berthelot2019remixmatch} enhance the MixMatch approach by incorporating two techniques: distribution alignment and augmentation anchoring. Recently, SimMatch \citeblue{zheng2022simmatch} has been introduced, where consistency regularization is applied at both the semantic level and instance level. This encourages the augmented views of the same instance to have consistent class predictions and similar relationships with respect to other instances. Additionally, \citetblue{lee2022contrastive} propose contrastive regularization to enhance the efficiency and accuracy of consistency regularization by leveraging well-clustered features of unlabeled data. \citetblue{verma2022interpolation} propose a straightforward and computationally efficient approach called Interpolation Consistency Training (ICT) for training deep neural networks in the SSL paradigm.

Second, various approaches \citeblue{sohn2020fixmatch, cascante2021curriculum, zhang2021flexmatch, kim2021distribution} provide sample selection strategies to generate pseudo-labels. For instance, FixMatch \citeblue{sohn2020fixmatch} combines consistency regularization and pseudo-labeling to obtain optimal performance and selects highly confident predictions as pseudo-labels using a predefined threshold. Instead of using a fixed threshold, \citetblue{zhang2021flexmatch} propose a method called Flexmatch, which dynamically adjusts thresholds for different classes at each time step. This allows informative unlabeled data and their pseudo-labels to be included. However, Flexmatch does not specifically address scenarios involving data imbalance. To tackle this particular issue, \citetblue{kim2021distribution} propose Distribution Aligning Refinery of Pseudo-label (DARP). It solves the various class-imbalanced SSL scenarios. The co-learning framework (CoSSL) \citeblue{fan2022cossl} addresses imbalanced SSL through decoupled representation learning and classifier learning.

The use of pseudo-labeling for multi-class classification problems presents a challenge in selecting accurate pseudo-labeled samples. Moreover, accurately estimating a class-wise threshold that accounts for imbalanced learning and correlations between classes would enable more accurate pseudo-label predictions. However, such a class-wise threshold is hard to estimate. To overcome these challenges and improve the reliability and accuracy of pseudo-labeling, we propose a novel approach that incorporates a class-wise prototype to identify similar unlabeled samples and perform pseudo-labeling using a similarity classifier, a KNN classifier, and a linear classifier.

\subsection{Semi-supervised learning in medical imaging}

Application of SSL methods in medical image analysis is an active field of research  \citeblue{van2020survey, hussain2022active, huang2023semi, lu2023boundary, farooq2023residual}. An effective SSL method over medical images can help in decreasing misdiagnosis rates significantly. Below is a review of recent SSL methods that have been applied in the field of medical imaging. \\
\textbf{Adversarial learning methods:} In medical image analysis, some studies have investigated SSL methods based on \textit{generative adversarial networks (GANs) \citeblue{goodfellow2020generative}}, demonstrating their broad applicability for automated diagnosis of heart \citeblue{madani2018deep, madani2018semi}, and retina disease \citeblue{lecouat2018semi, diaz2019retinal, wang2021deep}. For example, in \citetblue{madani2018semi}, GAN is utilised to overcome labelled data scarcity and data domain variance in the categorization of chest X-rays. In \citetblue{lecouat2018semi}, a semi-supervised GANs-based framework for patch-based classification is introduced. GANs are utilized for automated glaucoma assessment in the study by \citetblue{diaz2019retinal}.  \citetblue{wang2021deep} create adversarial samples using the virtual adversarial training technique in an effort to successfully explore the decision boundary. On the other hand, \citetblue{li2020shape} incorporate adversarial learning to leverage the shape information present in unlabeled data, promoting close proximity between the signed distance maps derived from labeled and unlabeled predictions.\\

{\textbf{Consistency-based methods: } Our study is  related to this line of work, which is widely employed for SSL {\citeblue{li2018semi, laine2016temporal, gyawali2020semi, wang2021neighbor}. Consistency-based methods ensure that predictions remain consistent across different augmentations of the same image. The predictions generated on augmented samples, known as consistency targets, play a critical role in the effectiveness of these approaches. It is essential to establish high-quality consistency targets during training to achieve optimal performance. The Pi model \citeblue{li2018semi} directly utilizes the network outputs as the consistency targets. The paper by \citetblue{wang2023deep} introduces deep semi-supervised multiple instance learning with self-correction. In this approach, a pseudo-label is generated for a weakly augmented input only if the model is highly confident in its prediction. This pseudo-label is then used to supervise the same input in a strongly augmented version. On the other hand, FullMatch, proposed by \citetblue{articleBCCD}, incorporates adaptive threshold pseudo-labeling to dynamically modify class thresholds according to the model's learning progress during training.

Other methods make use of ensembling information from previous epochs to calculate consistency targets. For instance, the Temporal Ensembling (TE) method \citeblue{laine2016temporal} defines consistency targets as Exponential Moving Average (EMA) predictions on unlabeled data. However, such a method has large memory requirements during training. GLM \citeblue{gyawali2020semi} produces enhanced samples by using mixup in both sample and manifold levels and minimise the distance across them. Neighbour matching (NM) \citeblue{wang2021neighbor} reweights pseudo-labels by using a feature similarity-based attention technique to align neighbour examples from a dynamic memory queue. Mean Teacher (MT) \citeblue{tarvainen2017mean} builds a teacher model using EMA for the model’s parameters. The prediction from the resulting teacher model is then used as the consistency targets for the original model. Based on MT, Local-teacher \citeblue{su2019local} incorporates a label propagation (LP) step, where a graph is constructed using the LP predictions, capturing both local and global data structure. To learn local and global consistency from the graph, a Siamese loss is employed. In order to encourage the model to uncover more semantic information from unlabeled data, SRC-MT \citeblue{liu2020semi} explicitly enforces the consistency of semantic relation among several samples under perturbations.

Unlike MT, which uses a temporal ensemble to update a teacher network, noteacher \citeblue{unnikrishnan2021semi} leverages two separate networks, eliminating the requirement for a teacher network. Within the field of medical imaging, MT has been widely used and adapted for segmentation tasks \citeblue{perone2018deep, yu2019uncertainty, hu2022semi}. Specifically, \citetblue{yu2019uncertainty} introduce a variation called Uncertainty-Aware Mean Teacher (UA-MT), where an uncertainty map is utilized to enhance the predictions of the teacher model. Another study by \citetblue{hu2022semi} incorporates uncertainty estimation to weigh the predictions of the mean teacher model, ensuring better nasopharyngeal carcinoma segmentation. MT {\citeblue{tarvainen2017mean} demonstrates an advantage over supervised learning when the teacher model produces better expected targets or pseudo-labels to train the student model. However, since the teacher model is essentially a temporal ensemble of the student model in the parameter space, MT is susceptible to confirmation bias or unintended propagation of label noise  \citeblue{ke2019dual, pham2021meta}. \\\\

\textbf{Other SSL methods:} In addition to the categories mentioned earlier, another approach employed by some techniques is multi-task learning, which is a widely utilized strategy for simultaneously learning multiple related tasks. The goal of multi-task learning is to leverage knowledge from one task to benefit others, ultimately improving generalizability \citeblue{zhang2021survey}. For example, \citetblue{gao2023semi} apply semi-supervised multi-task learning to weakly annotated whole-slide images, while \citetblue{wang2022semi} incorporate multi-task learning and contrastive learning into mean teacher to enhance the feature representation. The Anti-Curriculum Pseudo-Labeling (ACPL) approach \citeblue{liu2022acpl} employs a unique mechanism for selecting informative unlabeled samples and estimating pseudo-labels using mix-type classifiers without relying on a fixed threshold.

Our method is related to consistency-based SSL methods, which aim to enforce the model's predictions to be consistent across different augmentations of the same image. However, we specifically design weak and strong augmentations suitable for medical images to avoid distortion of critical features that differentiate one disease from another. Furthermore, Our method preserves a prototype for each class generated from a subset of the most recently observed training samples for that class. In order to identify reliable unlabeled samples, SPLAL measures their similarity with the class prototypes. Pseudo-labels for these reliable unlabeled samples are determined using a similarity classifier, a KNN classifier, and a linear classifier.

\section{Methods}
\label{sec:methods}

To introduce our SPLAL method, let us assume that we have a small labelled training set $\mathcal{D}_L = \{ (\mathbf{x}_i,\mathbf{y}_i)\}_{i=1}^{|\mathcal{D}_L|}$, where $\mathbf{x}_i \in \mathcal{X} \subset \mathbb{R}^{H \times W \times C}$ is the input image of size $H \times W$ with $C$ colour channels, and $\mathbf{y}_i  \in \{0,1\}^{\mathcal{K}}$ is the label. Here, $\mathcal{K}$ is total number of classes, and $\mathbf{y}_i$ is a one-hot vector. Let us say that we have a large unlabeled training set $\mathcal{D}_U = \{ \mathbf{x}_i\}_{i=1}^{|\mathcal{D}_U|}$, with $|\mathcal{D}_L| << |\mathcal{D}_U|$. We assume that the samples from both datasets are drawn from the same (latent) distribution. Our approach aims to learn a model $p:\mathcal{X} \to [0,1]^{\mathcal{K}}$, using only the labeled and pseudo-labeled samples. Let us define the model $p_{(\theta, \phi)}(.)$ as follows:
\begin{equation}
\begin{split}
\label{eq:model}
p = f_\phi^\text{cls} \circ f^{\text{enc}}_\theta
\end{split}
\end{equation}

In other words, $p_{(\theta, \phi)}(.)$ consists of a feature encoder $f^{\text{enc}}_\theta$ followed by a linear classifier $f_\phi^\text{cls}$. Here, $\theta$ and $\phi$ are the set of parameters of $f^{\text{enc}}_\theta$ and $f_\phi^\text{cls}$ respectively.

\begin{algorithm}
  \caption{SPLAL Algorithm}
  \begin{algorithmic}[1]
  \label{alg:guavaApproach}
    \STATE \textbf{require: } Labelled set $\mathcal{D}_L$, unlabeled set $\mathcal{D}_U$, and number of training stages $T$
    \STATE \textbf{warm-up train} {$p_{(\theta, \phi)}(.)$} using $\mathcal{L}$ as in Eq.~\eqref{eq:eqL1}.
    \STATE \textbf{initialise} set of class prototypes $\mathcal{C} = \{c_i\}  \text{ }\forall \text{ }k \in \{1,...,\mathcal{K}\}$ using a portion of recently viewed samples for training and $t=0$. 
    \WHILE{$t < T$ and $|\mathcal{D}_U| \neq 0$}
    \STATE \textbf{initialize} an empty set $\mathcal{D}_R$
    \STATE \textbf{select set of reliable samples} $\mathcal{D}_R$ \\ {$\mathcal{D}_R = 
        \{\mathbf{x}_u : \mathbf{x} \in \mathcal{D}_U,  h(\mathcal{C}, \mathbf{x}_u) = 1\}$}\\\text{as defined in Sec.~\eqref{sec:prototype}}
    \STATE \textbf{Estimate pseudo-label of $\mathbf{x}_u \in \mathcal{D}_R$} using Eq.~\eqref{eq:pseudo_label_ensemble} and Eq.~\eqref{eq:pseudo_label_weighted_prediction}
    \STATE \textbf{update labelled and unlabeled sets:} \\ {$\mathcal{D}_L \leftarrow \mathcal{D}_L \bigcup \mathcal{D}_R, \mathcal{D}_U \leftarrow \mathcal{D}_U \setminus \mathcal{D}_R$}
    \STATE \textbf{optimize} $\mathcal{L}$ as defined in Eq.~\eqref{eq:eqL1} using {$\mathcal{D}_L$ to obtain  $p_{(\theta, \phi)}(.)$}
    \STATE \textbf{update set of class prototypes} $\mathcal{C}$ using a portion of recently viewed samples for training. 
    \STATE {$t \leftarrow t + 1$}
    \ENDWHILE
    \RETURN $p_{(\theta, \phi)}(.)$
  \end{algorithmic}
\end{algorithm}

Our complete approach is described in Alg.~\myref{alg:guavaApproach}. In Sec.~\myref{sec:initial_training}, we introduce mathematical formulations of the alignment loss used in our loss function. In Sec.~\myref{sec:prototype}, we describe our approach for reliable sample selections using class prototypes. In Sec.~\myref{sec:pseudo_labeling}, we introduce our pseudo-labeling procedure in detail. We use a weighted combination of a similarity classifier, a KNN classifier and a linear classifier to predict pseudo-labels. Fig.~\myref{fig:loss} gives a pictorial representation of our approach SPLAL.

\subsection{SPLAL optimization}
\label{sec:initial_training}
Our SPLAL optimization, described in Alg.~\myref{alg:guavaApproach} and depicted by Fig.~\myref{fig:splal optim}, starts with a warm-up supervised training of the model $p_{(\theta, \phi)}(.)$ using only the labeled set $\mathcal{D}_L$. For subsequent training, we use the updated labeled dataset, which also contains the pseudo-labeled samples. In every iteration, we try to minimize the loss function given in {Eq.}~\eqref{eq:eqL1}. It includes a classification loss and an alignment loss. Alignment loss is used to  enforce the model to give similar predictions for different augmentations of the same image. Thus, we generate two augmentations for every image -- weak and strong. Our aim is to learn the model $p_{(\theta, \phi)}(.)$, such that, along with the classification loss, the difference between the model's prediction for the two augmentations of the same image is also minimized. We should consider the choice of augmentation carefully to prevent any significant change in the distinguishing feature of the corresponding class, as it can lead to the misclassification of the image to other classes.

Let $\mathbf{x}_i$ be a labeled sample. We define $\mathcal{L}$ as the total loss function of our model $p_{(\theta, \phi)}(.)$, by

\begin{equation}
\begin{split}
\label{eq:eqL1}
\mathcal{L}(\theta, \phi, \mathcal{D}_L) & =\lambda_{1}\sum_{(\mathbf{x}_i,\mathbf{y}_i)  \in \mathcal{D}_L}
\ell(\mathbf{y}_i,p_{(\theta, \phi)}(\mathbf{x}_i))\\
& + \lambda_{2}\sum_{(\mathbf{x}_i,\mathbf{y}_i)  \in \mathcal{D}_L}
\ell({\hat{\mathbf{y}}^{weak}}_{i},{\hat{\mathbf{y}}^{strong}}_{i})%
\end{split}
\end{equation}

Here, $\ell(.)$ denotes a standard loss function (\textit{e.g}., cross-entropy loss) and $\mathbf{y}_i$ is the ground truth. ${\hat{\mathbf{y}}^{weak}}_{i}$ and ${\hat{\mathbf{y}}^{strong}}_{i}$ represent the prediction of our model for weak augmentation (${\mathbf{x}}^{weak}_{i}$) and strong augmentation (${\mathbf{x}}^{strong}_{i}$) of $\mathbf{x}_i$, respectively. $\lambda_{1}$ and $\lambda_{2}$ are hyperparameters, such that $
\lambda_{1} + \lambda_{2} = 1$.

\subsection{Reliable sample selection}
\label{sec:prototype}
Our SPLAL approach, selects a set of reliable unlabeled samples based on their similarity with the class prototypes. Class imbalanced data can lead to the generation of imbalanced prototypes. To avoid this, we use a prototype generation framework inspired by DASO \citeblue{oh2022daso}. A memory queue dictionary $\mathbf{Q}=\{Q_{k}\}^{\mathcal{K}}_{k=1}$ is maintained, where each key represents a class, and $Q_k$ refers to the memory queue for class $k$. The size of the memory queue for each class is kept constant. Memory queue $Q_k$ is updated for each class $k$ at every training iteration by pushing new features from labeled data in the batch and removing the oldest ones when $Q_k$ is full. The class prototype $c_k$ is computed for each class $k$ by averaging the feature vectors in the queue $Q_k$.

We define $\mathcal{C}$ as the set of all class prototypes $\mathbf{c}_k$, for each class $\mathbf{k} \in \{1,...,\mathcal{K}\}$. Let $h(\mathcal{C}, x_u)$ be a function that measures the reliability of unlabeled sample $x_u$ as follows:
\begin{equation}
    h(\mathcal{C}, x_u) =
\left\{\begin{array}{ll} 
1, & \text{if $x_u$ is reliable},\\
0, & \text{otherwise}
\end{array}
\right.	
    \label{eq:reliability_estimator}
\end{equation}
Let $w$ be a vector of size $\mathcal{K}\times1$. Let $w_k$ be the value of the \textit{k\textsuperscript{th}} row in vector $w$, where $1 \leq k \leq \mathcal{K}$. Then, $w_k$ is defined by,

\begin{equation}
\label{eqn:cosineSimilarityVector}
w_k = \mathbf{sim}(c_{k}, f^{\text{enc}}_\theta(x_u)) \;\; \forall \;\; k \in \{1,...,\mathcal{K}\}
\end{equation}
Here, $f^{\text{enc}}_\theta(x_u)$ gives the feature vector for $x_u$, and $\textbf{sim}$ represents cosine similarity function. Let us define a vector $v$ by passing $w$ to the softmax function as described below:
\begin{equation}
\label{eqn:cosineSimilaritySoftmax}
v = \sigma(w)
\end{equation}
Here, $\sigma$ denotes the softmax function. Let $v_k$ be the value of \textit{k\textsuperscript{th}} row in vector $v$, where $1 \leq k \leq \mathcal{K}$. Value of $h(\mathcal{C}, x_u)$ is 1, iff, $\exists$ an index $j$, such that $v_j$ $\geq \gamma_1$; and $v_i \leq \gamma_2 \; \forall \; i \in \{1,...,\mathcal{K}\} \setminus j$, otherwise 0.

Here, $\gamma_1$ and $\gamma_2$ are hyperparameters. It is worth noting that higher the value of $\gamma_1$, the higher the reliability of the samples selected. However, less number of samples will be selected. On the other hand, the lower the value of the constant $\gamma_1$, the higher the number of unlabeled samples selected for pseudo-labeling. However, the reliability of the selected samples can not be confidently guaranteed.

\renewcommand{\arraystretch}{1.2}
\begin{table*}[!t]
\centering
\caption{Analysis of AUC, specificity, accuracy, and F1 score of state-of-the-art SSL methods on skin lesion classification (ISIC 2018) dataset with 20$\%$ labeled data. The best result under each evaluation metric is highlighted in bold. Here, * denotes the result using DenseNet-169 as the backbone model and $^\dagger$ represents that the results are taken from FullMatch \citeblue{articleBCCD}.}
\label{table:isic18_main}
\scalebox{0.90}{
    \begin{tabular}{l l l l l l l}
        \hline
        \multicolumn{1}{l}{Method} & \multicolumn{2}{c}{Percentage} & \multicolumn{4}{c}{Metrics} \\
        
        \cmidrule(r){2-3}
        \cmidrule(r){4-7}
        
          & Labeled & Unlabeled & AUC & Specificity & Accuracy & F1 score\\
        \hline
             Baseline$^\dagger$ & 20\% & 0 & 90.15 & 91.83 & 92.17 & 52.03\\
             Self-training$^\dagger$ \citeblue{bai2017semi} & 20\% & 80\% & 90.58 & 93.31 & 92.37 & 54.51\\
             SS-DCGAN$^\dagger$ \citeblue{diaz2019retinal} & 20\% & 80\% & 91.28  & 92.56 & 92.27 & 54.10\\
             TCSE$^\dagger$ \citeblue{li2018semi} & 20\% & 80\% & 92.24  & 92.51 & 92.35 & 58.44\\
             TE$^\dagger$ \citeblue{laine2016temporal} & 20\% & 80\% & 92.70 & 92.55 & 92.26 & 59.33\\
             MT$^\dagger$ \citeblue{tarvainen2017mean} & 20\% & 80\% & 92.96 & 92.20 & 92.48 & 59.10\\
             SRC-MT$^\dagger$ \citeblue{liu2020semi} & 20\% & 80\% & 93.58  & 92.72 & 92.54 & 60.68\\
             FixMatch$^\dagger$ \citeblue{sohn2020fixmatch} & 20\% & 80\% & 93.83 & 92.18 & 93.39 & 61.64\\
             FixMatch$+$DARP$^\dagger$ \citeblue{kim2021distribution} & 20\% &80\% & 94.02 & 92.46 & 93.43 & 62.05\\
             FlexMatch $^\dagger$\citeblue{zhang2021flexmatch}& 20\% & 80\% & 93.55 & 92.32 & 93.41 & 60.90\\
             ACPL$^\dagger$ \citeblue{liu2022acpl}& 20\% & 80\% & 94.36 & - & - & 62.23\\
             FullMatch$^\dagger$ \citeblue{articleBCCD} & 20\% & 80\% & 94.95 & 91.87 & 93.45 & 63.25\\
             Ours & 20\% & 80\% & \textbf{95.79} & \textbf{95.36} & \textbf{95.54} & \textbf{70.46}\\
             Ours* & 20\% & 80\% & \textbf{96.38} & \textbf{96.01} & \textbf{95.72} & \textbf{73.16}\\
        \hline
    \end{tabular}
    }

\end{table*}

\begin{figure*}[hbt!]
    \centering
    \scalebox{0.99}{
    \includegraphics[width=\linewidth]{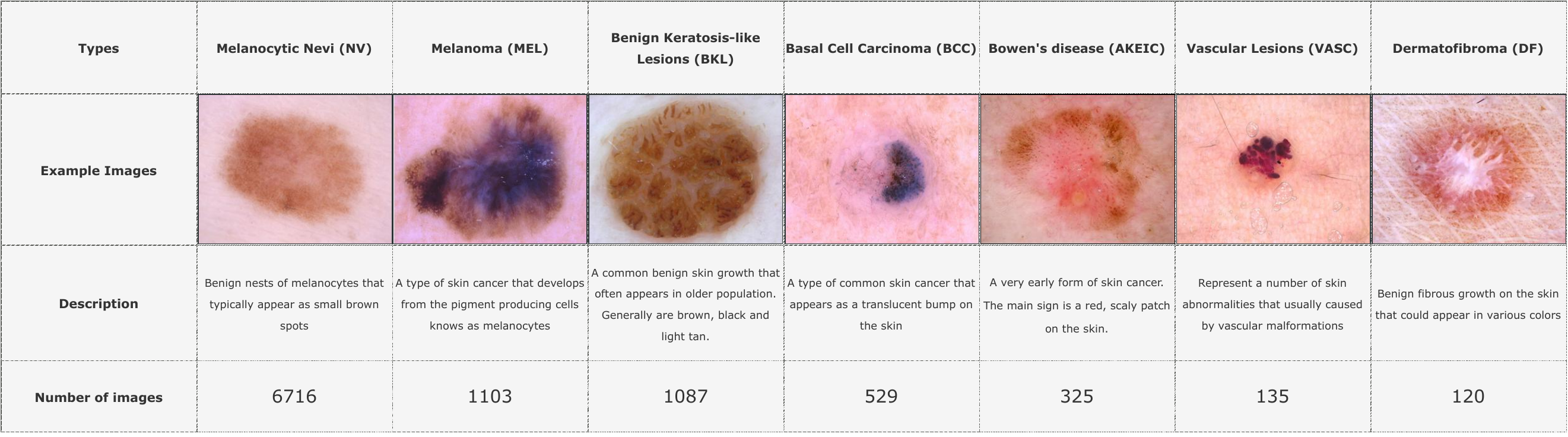}
    }
    \caption{Example images along with their detailed information from the ISIC 2018 dataset.}
    \label{fig:isic}
\end{figure*}

\subsection{SPLAL pseudo-labeling framework}
\label{sec:pseudo_labeling}
Once the set of reliable unlabeled samples is selected, we need to predict their accurate pseudo-labels so that the model can be effectively trained on those samples. Our approach uses a weighted combination of a similarity classifier, a KNN classifier and a linear classifier for predicting pseudo-labels. The similarity classifier uses similarity with the class prototypes as a criterion for prediction. In KNN, we consider the $K$ closest samples from the updated labeled dataset for prediction. It must be noted that the updated labeled dataset contains both the labeled and the pseudo-labeled samples.
 
Let us consider $\mathcal{D}_R$ to be the set of reliable samples selected by $h(.)$. Now for an unlabeled sample $x_u \in \mathcal{D}_R$, we define

\begin{equation}
    \label{eq:pseudo_label_ensemble}
    \begin{split}
        {\hat{\mathbf{y}}^{\text{linear classifier}}}_{u} &= p_{(\theta, \phi)}(\mathbf{x}_u), \\
         {\hat{\mathbf{y}}^{\text{KNN classifier}}}_{u} &= \frac{1}{K}\sum_{(f^{\text{enc}}_\theta(\mathbf{x}),\mathbf{y}) \in  \mathcal{N}(f^{\text{enc}}_\theta(\mathbf{x_u}),\mathcal{D}_L)}  \mathbf{y}, \\
         {\hat{\mathbf{y}}^{\text{similarity classifier}}}_{u} &= \mathbf{}{OneHot}(\arg \max_{1 \leq k \leq \mathcal{K}}(v_k))
    \end{split}
\end{equation}
Here, ${\hat{\mathbf{y}}^{\text{linear classifier}}}_{u}, {\hat{\mathbf{y}}^{\text{KNN classifier}}}_{u}$, and ${\hat{\mathbf{y}}^{\text{similarity classifier}}}_{u}$ represent the prediction of the respective classifiers over $x_u$ and $\mathcal{N}(f^{\text{enc}}_\theta(\mathbf{x_u}),\mathcal{D}_L)$ represents the set of K-nearest neighbors from
the labeled set $\mathcal{D}_L$ to the $f^{\text{enc}}_\theta(\mathbf{x_u})$, with each element in the set $\mathcal{D}_L$ denoted by $(f^{\text{enc}}_\theta(\mathbf{x}),\mathbf{y})$. The final pseudo-label of $x_u$ is given by: 

\begin{equation}
\begin{split}
    {\hat{\mathbf{y}}^{\text{pseudo-label}}}_{u}  & = \alpha_{1}\times{\hat{\mathbf{y}}^{\text{linear classifier}}}_{u} \\ 
    & \quad+ \alpha_{2}\times{\hat{\mathbf{y}}^{\text{KNN classifier}}}_{u} \\
    & \quad+ \alpha_{3}\times{\hat{\mathbf{y}}^{\text{similarity classifier}}}_{u}\\
    \label{eq:pseudo_label_weighted_prediction}
\end{split}
\end{equation}

Here, ${\hat{\mathbf{y}}^{\text{pseudo-label}}}_{u}$ is the estimated pseudo-label of $x_u$, and $\alpha_{1}$, $\alpha_{2}$ and $\alpha_{3}$ are the hyperparameters, such that $\alpha_{1} + \alpha_{2} + \alpha_{3} = 1$. The importance of $\alpha_{1}$, $\alpha_{2}$, and $\alpha_{3}$ is worth noting. Since, similarity with class prototypes is a necessary criterion for reliable sample selection, $\alpha_{3}$ has a comparatively higher value (as justified in Sec ~\myref{sec:WeightOfClassifierForPL}), which implies that the prediction of the similarity classifier dominates in pseudo-label prediction. However, $\alpha_{1}$ and $\alpha_{2}$ allow label smoothing by accounting for the prediction of the KNN classifier and the linear classifier. Once the reliable unlabeled samples are selected and their pseudo-label estimated, we add them to the labeled dataset and remove them from the unlabeled dataset. Thus, after pseudo-labeling, the labeled and unlabeled sets are updated as $\mathcal{D}_L = \mathcal{D}_L \bigcup \mathcal{D}_R$, and $\mathcal{D}_U = \mathcal{D}_U \setminus \mathcal{D}_R$, and the next iteration of optimization and updation of class prototypes takes place.
\section{Experiments and results}
\label{sec:experiments}
To evaluate the effectiveness of our proposed approach, we perform extensive experiments on two publicly available datasets: the skin lesion classification (ISIC 2018) \citeblue{tschandl2018ham10000, codella2019skin} dataset and the blood cell classification dataset (BCCD) \citeblue{bloodCellImages}. Additionally, we analyze the performance of our approach across varying ratios of labeled data for the ISIC 2018 dataset. We also perform comprehensive ablation studies on the ISIC 2018 dataset to validate the contribution of different components of our approach. The results of our experiments are presented in Sec.\myref{sec:results}, while the details of the ablations are discussed in Sec.\myref{sec:ablation_study}.

\subsection{Datasets and experimental setup}
\label{sec:dataset_and_implementation}

\subsubsection{Skin lesion classification dataset (ISIC 2018)}
\label{sec:isic2018}
The ISIC 2018 \citeblue{tschandl2018ham10000, codella2019skin} is a skin lesion challenge dataset organized by the International Skin Imaging Collaboration (ISIC). It has a highly imbalanced distribution, as depicted by Fig.~\myref{fig:isic18_distribution}. It contains 10,015 images with seven labels. Each image is associated with one of the labels in -- Melanocytic Nevi, Melanoma, Benign Keratosis-like Lesions, Basal Cell Carcinoma, Bowen's disease, Vascular Lesions, and Dermatofibroma. An overview of details of each disease, along with the example image, is presented in Fig.~\myref{fig:isic}. Out of the total number of images, we consider 80$\%$ as training images and 20$\%$ testing images. For the train/test split, we follow the division as given in \citeblue{liu2020semi}.

\subsubsection{Blood cell classification dataset (BCCD)}
\label{sec:bccd_Kaggle}
BCCD \citeblue{bloodCellImages} is a blood cell classification dataset publicly available over the Kaggle platform. It has a relatively balanced distribution, as depicted by Fig.~\myref{fig:bccd_distribution}. It contains 12,442 augmented blood cell images with four labels. Each image is associated with one of the labels in -- Eosinophils, Lymphocytes, Monocytes, and Neutrophils. An overview of details of each disease, along with the example image, is presented in Fig.~\myref{fig:bccd}. For experiments, we keep the division of the original dataset \citeblue{bloodCellImages}, and remove duplicate images from the training and testing datasets. There are 9898 images in the training dataset and 2465 images in the test dataset, each with only one label.

\begin{figure}[hbt!]
    \centering
    \scalebox{0.99}{
    \includegraphics[width=\linewidth]{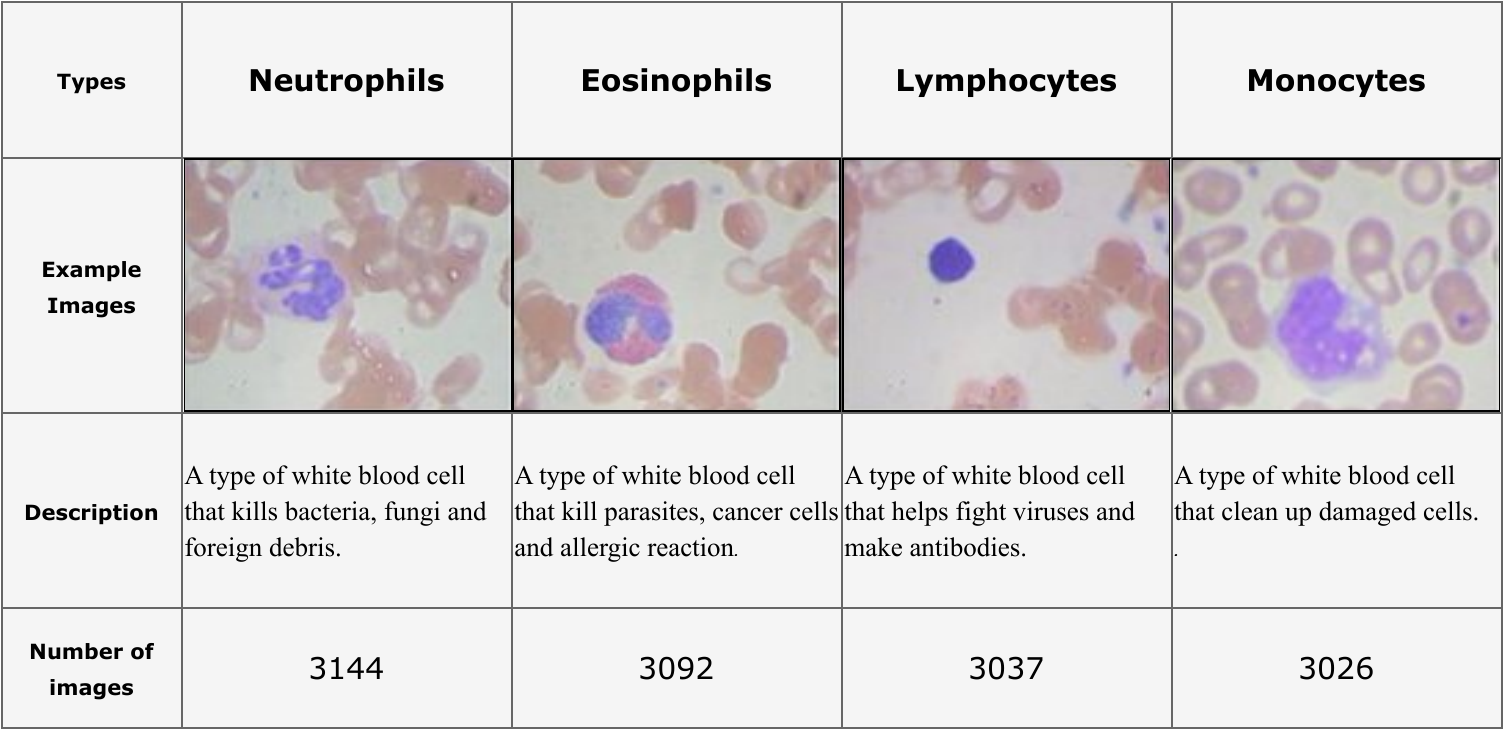}
    }
    \caption{Example images along with their detailed information from the blood cell classification dataset (BCCD).}
    \label{fig:bccd}
\end{figure}

\begin{table*}[h!]
\centering
\caption{Analysis of accuracy, sensitivity, specificity, precision, and F1 score of several state-of-the-art SSL methods on the blood cell classification dataset (BCCD) with 20$\%$ labeled data. The best result under each evaluation metric is highlighted in bold. Here, * denotes the result using DenseNet-169 as the backbone model and $^\dagger$ represents that the results are taken from FullMatch \citeblue{articleBCCD}.}
\label{table:bccd_main}
\scalebox{0.90}{
    \begin{tabular}{l l l l l l l l}
        \hline
        \multicolumn{1}{l}{Method} & \multicolumn{2}{c}{Percentage} & \multicolumn{5}{c}{Metrics} \\
        
        \cmidrule(r){2-3}
        \cmidrule(r){4-8}
        
          & Labeled & Unlabeled & Accuracy & Sensitivity & Specificity & Precision & F1 score \\
        \hline
             Baseline$^\dagger$ & 20\% & 0 & 91.08 & 85.95 & 92.80 & 81.52 & 82.95\\
             MT$^\dagger$ \citeblue{tarvainen2017mean} & 20\% & 80\% & 94.42 & 90.53 & 95.71 & 89.46 & 89.22\\
             SRC-MT$^\dagger$ \citeblue{liu2020semi} & 20\% & 80\% & 94.57 & 90.58 & 95.88 & 90.02 & 89.49\\
             FixMatch$^\dagger$ \citeblue{sohn2020fixmatch} & 20\% & 80\% & 94.24 & 89.84 & 95.69 & 89.97 & 88.97\\
             FixMatch$+$DARP$^\dagger$ \citeblue{kim2021distribution} & 20\% &80\% & 94.56 & 90.60 & 95.87 & 90.38 & 89.55\\
             FlexMatch$^\dagger$ \citeblue{zhang2021flexmatch} & 20\% & 80\% & 94.50 & 90.46 & 95.84 & 90.22 & 89.43\\
             FullMatch$^\dagger$ \citeblue{articleBCCD} & 20\% & 80\% & 94.88 & \textbf{90.61} & 96.29 & 91.25 & 90.04\\
             Ours & 20\% & 80\% & \textbf{95.13} & 90.24 & \textbf{96.74} & \textbf{92.10} & \textbf{90.41} \\
             Ours* & 20\% & 80\% & \textbf{95.25} & 90.44 & \textbf{96.81} & \textbf{92.19} & \textbf{90.66} \\
        \hline
    \end{tabular}
}
\end{table*}

\subsubsection{Experimental details}
\label{sec:implementation_details}
We train our model with a Tesla RTX A5000. For both datasets, we use DenseNet-121 \citeblue{huang2017densely}, pre-trained on ImageNet \citeblue{russakovsky2015imagenet} as our backbone model. We use Adam \citeblue{kingma2017adam} optimizer for training. The batch size is 32 and 16 for the ISIC 2018 and BCCD, respectively. The initial learning rate is 0.03 and 0.009 for ISIC 2018 and BCCD, respectively. The image size of both datasets is adjusted to 224 $\times$ 224 for faster processing. For both datasets, we perform 50 epochs for warm-up training and an additional 40 epochs whenever we mix a reliable set of unlabeled samples with the labeled dataset. The values of $\alpha_1$, $\alpha_2$, and $\alpha_3$ are 0.20, 0.10, and 0.70 for both datasets. The value of $\gamma_1$ is 0.99 and 0.90 for the ISIC 2018 and BCCD, respectively. The value of $\gamma_2$ is 0.005 and 0.03 for the ISIC 2018 and BCCD, respectively. The value of $\lambda_1$ is 0.60 and 0.75 for the ISIC 2018 and BCCD, respectively. Consequently, the value of $\lambda_2$ is 0.40 and 0.25 for the ISIC 2018 and BCCD, respectively. The value of $K$ for the KNN classifier is 200 for both datasets. We use random horizontal and vertical flips for weak augmentation and Gaussian blur for strong augmentation. We use Pytorch\citeblue{NEURIPS2019_bdbca288} for our implementation. We maintain an exponential moving average (EMA) version of the trained model, as given in \citeblue{liu2021self, liu2020semi, tarvainen2017mean}. It is important to note that the EMA version of the model is used only for evaluation and not for training.

\subsection{Results}
\label{sec:results}

\subsubsection{Results of the skin lesion classification dataset}
\label{sec:results_isic18}
On the ISIC 2018 dataset, we compare our method to Self-training \citeblue{bai2017semi}, GAN-based method \citeblue{diaz2019retinal}, $\Pi$ model-based method \citeblue{li2018semi}, Temporal Ensembling (TE) \citeblue{laine2016temporal}, Mean Teacher (MT) \citeblue{tarvainen2017mean}, and SRC-MT \citeblue{liu2020semi}. In addition, we compare our method with some pseudo-labeling based SSL methods, namely -- FixMatch \citeblue{sohn2020fixmatch}, FlexMatch \citeblue{zhang2021flexmatch}, and FullMatch \citeblue{articleBCCD}. We also compare with ACPL \citeblue{liu2022acpl} and Distribution Aligning Refinery of Pseudo-label (DARP) \citeblue{kim2021distribution}, which are SSL methods to solve the imbalanced problem. The backbone model for all these methods is DenseNet-121 \citeblue{huang2017densely}. The performance of these methods on the ISIC18 dataset with 20$\%$ labeled data is summarized in Table~\myref{table:isic18_main}. It is worth noting that our approach achieves better results than other contemporary SSL methods in terms of AUC, accuracy, specificity, and F1 score.

\subsubsection{Results of the blood cell classification dataset}
\label{sec:results_bccd_Kaggle}

On the blood cell classification dataset, we compare our method with MT \citeblue{tarvainen2017mean}, SRC-MT \citeblue{liu2020semi}, FixMatch \citeblue{sohn2020fixmatch}, DARP \citeblue{kim2021distribution}, FlexMatch \citeblue{zhang2021flexmatch} and FullMatch \citeblue{articleBCCD}. The backbone model for all these methods is DenseNet-121 \citeblue{huang2017densely}. The performance of these methods on the blood cell classification dataset (BCCD) with 20$\%$ labeled data is shown in Table~\myref{table:bccd_main}. The results suggest that, except for sensitivity, our approach achieves better results than other contemporary SSL methods in every evaluation metric.

\subsection{Ablation studies}
\label{sec:ablation_study}

\begin{table}[t!]
    \centering
    \caption{Analysis of AUC and F1 score of our approach on the ISIC 2018 dataset, using varying ratio of labeled data. The comparison with the baseline method and FullMatch \citeblue{articleBCCD} is also described. The best result under each category is highlighted in bold.}
    \label{tab:varying_ratio_table}
    \scalebox{0.95}{
    \begin{tabular}{c|c|c|c|c}
    \toprule Method     & Label Ratio   & Accuracy & AUC & F1 score    \\ \hline
              Baseline & 5\% & 84.73 & 84.24       & 38.57 \\ 
              FullMatch  & 5\% & 89.82 & 90.66      & 50.64 \\ 
              Ours & 5\% & \textbf{93.58} & \textbf{92.80} & \textbf{55.37} \\ \hline
              
              Baseline & 10\% & 87.45 & 87.04       & 44.43 \\ 
              FullMatch & 10\% & 91.50 & 92.70      & 57.07 \\ 
              Ours & 10\% & \textbf{94.99} & \textbf{94.36} & \textbf{66.38} \\ \hline
              
              Baseline & 20\% & 92.17 & 90.15       & 52.03 \\ 
              FullMatch & 20\% & 93.45 & 94.95      & 63.25 \\ 
              Ours & 20\% & \textbf{95.54} & \textbf{95.79} & \textbf{70.46} \\ \hline
              
              Baseline & 30\% & 92.55 & 91.80       & 57.83 \\ 
              FullMatch & 30\% & 93.82 & 95.17      & 65.15 \\ 
              Ours & 30\% & \textbf{96.18} & \textbf{96.85} & \textbf{74.19} \\  \hline
              \bottomrule
    \end{tabular}}
  \end{table}

\begin{figure*}[h]
    \centering
    \scalebox{1}{
    \includegraphics[width=0.8\linewidth]{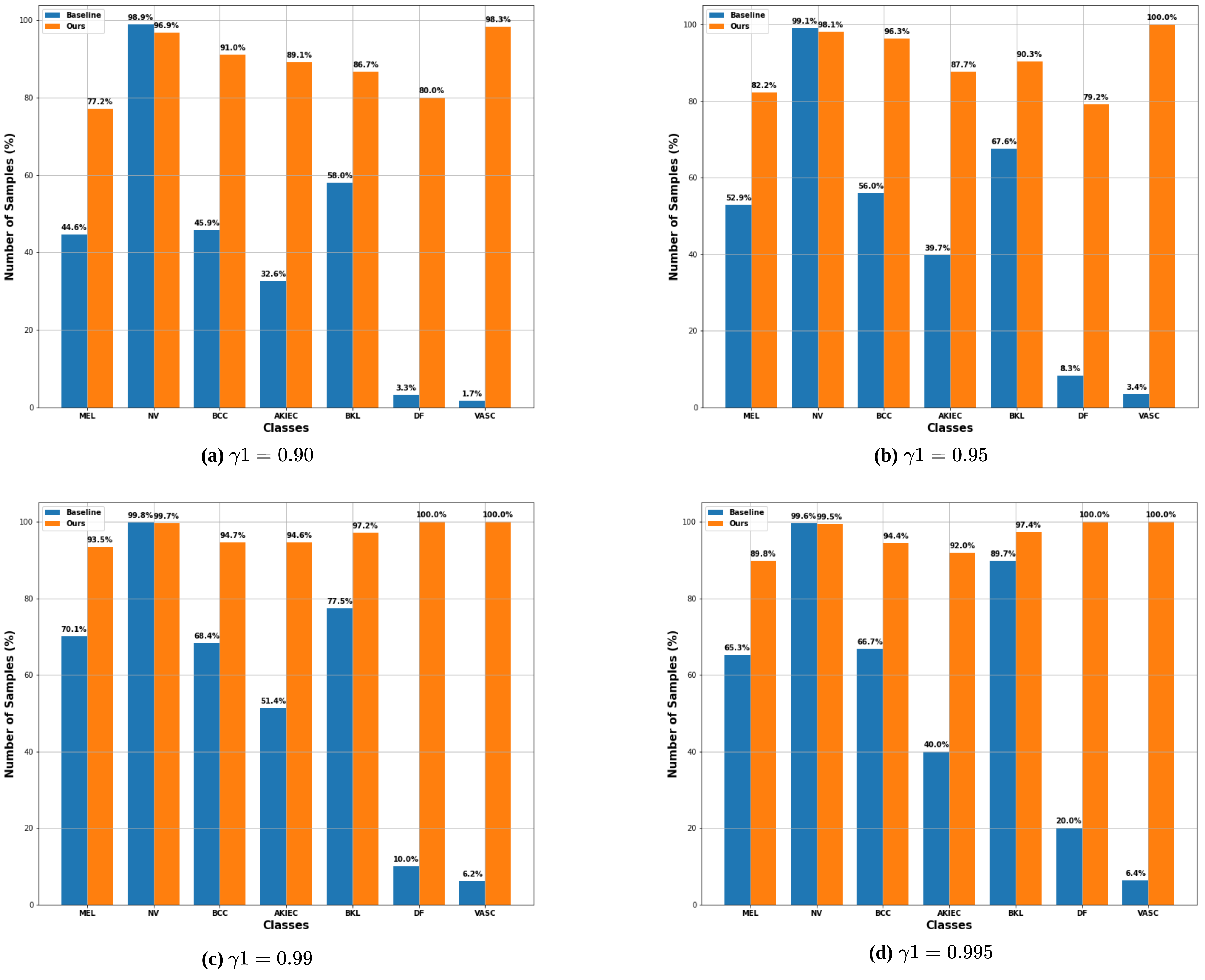}
    }
    \caption{Comparison between the percentage of pseudo-labels correctly predicted for a set of reliable unlabeled samples using the baseline method (in blue) and ours (in orange). The analysis is done across different values of $\gamma_1$.}
    \label{fig:reliability analysis Figure}
\end{figure*}

\subsubsection{Effect of different labeled data percentages on SPLAL}
\label{sec:different_label_ration}
We evaluate our SPLAL method on ISIC 2018 dataset with different percentages of labeled data, and the results are summarized in Table~\myref{tab:varying_ratio_table}. Our approach consistently outperforms the baseline and FullMatch \citeblue{articleBCCD} for the given evaluation metrics for all labeled data percentages. These results demonstrate that SPLAL can effectively leverage the information from unlabeled samples to improve classification performance, even with limited labeled data, highlighting our approach's robustness and generalizability.

\begin{table}[t!]
    \centering
    \caption{Analysis of accuracy, AUC, and F1 score of our approach on  the ISIC 2018 dataset, using different values of $\lambda_2$. The best result under each category is highlighted in bold. %
    }
    \label{tab:regularization_table}
    \scalebox{0.95}{
    \begin{tabular}{c|c|c|c}
    \toprule $\lambda_2$ & Accuracy & AUC  & F1 score    \\ \hline
              0.00 & 95.11 & 95.24 & 68.59 \\ \hline
              
              0.10 & 95.52 & 95.67 & 70.16 \\  \hline
              
              0.25 & 95.46 & 95.34 & 70.01 \\  \hline
              
              0.40 & \textbf{95.54} & \textbf{95.79} & \textbf{70.46} \\  \hline
              
              0.50 & 95.39 & 95.22 & 69.82 \\  \hline
              
              0.60 & 95.21 & 95.00 & 68.68 \\  \hline
              \bottomrule
    \end{tabular}}
\end{table}

\subsubsection{Effect of alignment loss in SPLAL optimization}
\label{sec:regulatization_effect}
The impact of $\lambda_{2}$ on SPLAL is described in Table~\myref{tab:regularization_table}. $\lambda_{2}$ essentially controls the weight of alignment loss in the total loss function. We infer that an appropriate value of $\lambda_{2}$ helps improve the classification performance for minority classes. As shown in Table~\myref{tab:regularization_table}, we achieve the best results in terms of accuracy, AUC, and F1 score, when $\lambda_{2}$ is 0.40. Due to $\lambda_{2}$, our approach gives similar predictions for different augmentation of an image, which helps maintain a consistent prediction for different samples belonging to minority classes. Fig.~\myref{fig:reliability analysis Figure} shows that the baseline method performs poorly on the minority classes. However, our approach performs significantly better on the minority classes, which can be attributed to the alignment loss.

\begin{table}[h!]
    \centering
    \caption{Analysis of accuracy, AUC, and F1 score of our approach on the ISIC 2018 dataset, using different values of $\gamma_1$. The best result under each category is highlighted in bold.}
    \label{table:ablation3}
    \scalebox{0.90}{
        \begin{tabular}{c|c|c|c}
        \toprule $\gamma_1$  & Accuracy & AUC  & F1 score \\ \hline
            0.90 & 95.31 & 94.82 & 69.48  \\ \hline
            0.95 & 95.36 & 95.14 & 69.63   \\ \hline
            0.99 & \textbf{95.54} & \textbf{95.79} & \textbf{70.46}  \\  \hline
            0.995 & 95.09 & 95.34  & 68.02  \\ \hline
            
         \bottomrule
        \end{tabular}
    }
\end{table}
    
\begin{table}[t!]
  \centering
  \caption{Analysis of accuracy, AUC, and F1 score of our approach on the ISIC 2018 dataset, using different combinations of classifiers for the estimation of pseudo-label. The best result under each category is highlighted in bold.}
  \label{tab:ensemble_pl_table}
  \scalebox{0.95}{
  \begin{tabular}{c|c|c|c}
  \toprule Combination of classifiers &    Accuracy & AUC & F1 score    \\ \hline
            Similarity + KNN + Linear & \textbf{95.54} & \textbf{95.79} & \textbf{70.46} \\ \hline
            
            Similarity + Linear & 95.38 & 95.69 & 69.05 \\  \hline
            
            Similarity + KNN & 95.36 & 95.40 & 69.99 \\  \hline
            \bottomrule
  \end{tabular}}
\end{table}

\begin{figure*}[h]
    \centering
    \scalebox{1}{
    \includegraphics[width=\linewidth]{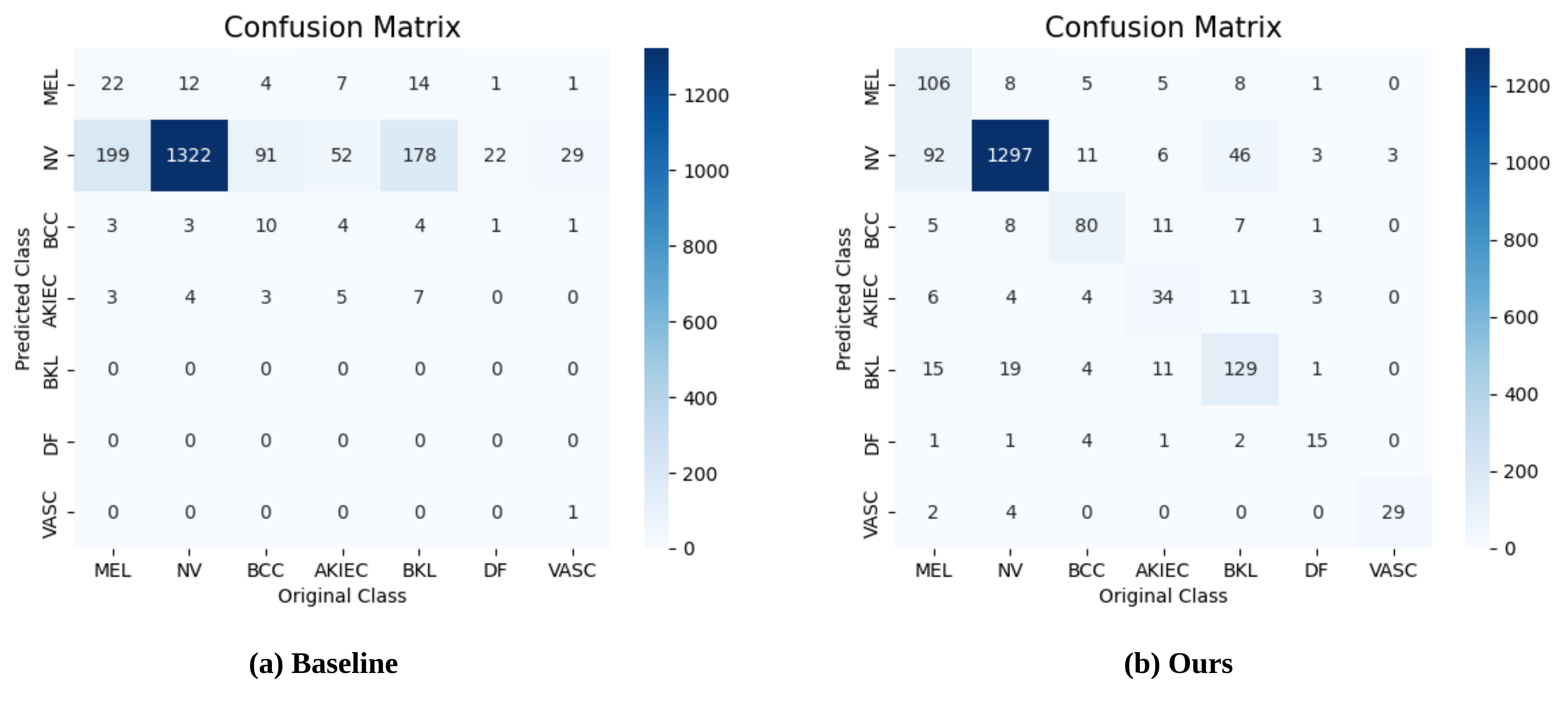}
    }
    \caption{Comparison between the confusion matrix generated by the baseline method and our approach. We can see that the classification of the baseline method is biased towards the majority class. However, our approach gives significant number of correct predictions in every class.}
    \label{fig:heatmap}
\end{figure*}

\begin{figure*}[h!]
    \centering
    \scalebox{0.95}{
    \includegraphics[width=\linewidth]{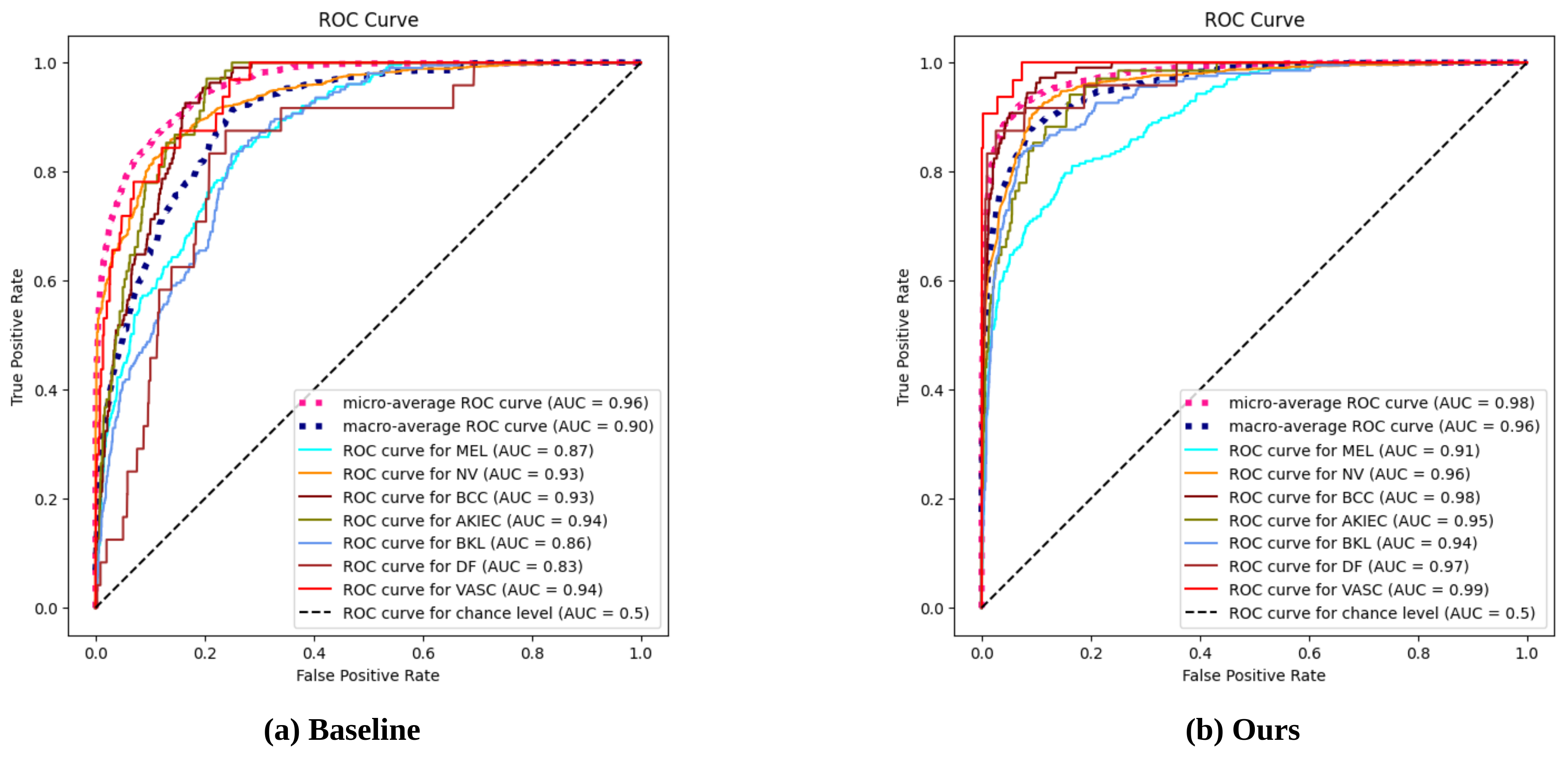}
    }
    \caption{Comparison between the ROC curve generated by the baseline method and our approach. We can see that our approach performs equally well for all the classes in terms of AUC as opposed to the baseline method.}
    \label{fig:rocCurveFigure}
\end{figure*}

\begin{figure*}[hbt!]
    \centering
    \scalebox{0.80}{
    \includegraphics[width=\linewidth]{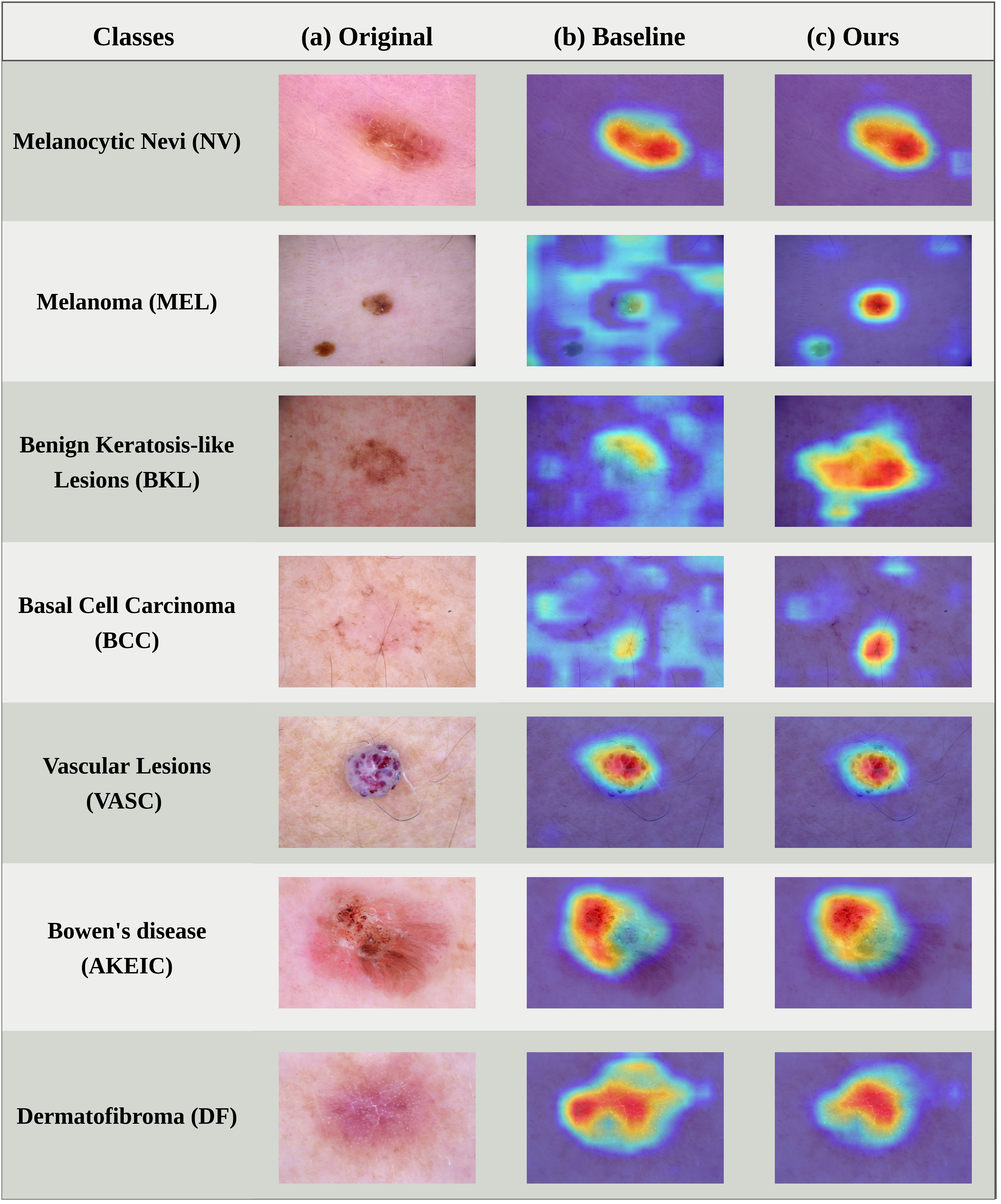}
    }
    \caption{Comparison between the Grad-CAM visualizations generated by the baseline method and our SPLAL method. We can see that the baseline method is not able to clearly use the distinguishing features of the disease for some of the classes. However, our approach gives prediction using the distinguishing feature of the disease for every class.}
    \label{fig:Gradcam}
\end{figure*}  

\subsubsection{Effect of $\gamma_1$ and $\gamma_2$ on reliable sample selection}
\label{sec:informationThreshold}
The hyperparameters $\gamma_1$ and $\gamma_2$ play an important role in reliable sample selection procedure. Table~\myref{table:ablation3} describes the impact of $\gamma_1$ and $\gamma_2$ on the performance in terms of various evaluation metrics. For analysis, we keep changing the value of $\gamma_1$ and keep the value of $\gamma_2$ as follows: 
\begin{equation}
\begin{split}
    \gamma_2 = \frac{|1-\gamma_1|}{2}
    \label{eq:relation between gamma2 and gamma1}
\end{split}
\end{equation}
Fig.~\myref{fig:reliability analysis Figure} shows the comparison between the percentage of correctly predicted pseudo-labels for a set of reliable unlabeled samples using our approach and the baseline method across different values of $\gamma_1$. The choice of $\gamma_1$ affects the set of reliable unlabeled samples selected. Higher the value of $\gamma_1$, higher will be the reliability of selected sample. Thus, higher the value of $\gamma_1$, higher will be the percentage of correctly predicted pseudo-labels by our approach. However, Table~\myref{table:ablation3} shows that the performance deteriorates if $\gamma_1$ is increased beyond a threshold. Thus, we infer that a value of $\gamma_1$, which is neither too high nor too low, is good for our method.

\begin{table}[h!]
    \centering
    \caption{Analysis of accuracy, AUC, and F1 score of our approach on the ISIC 2018 dataset, using different values of $\alpha_{1}$, $\alpha_{2}$, and $\alpha_{3}$. The best result under each category is highlighted in bold.}
    \label{table:ablation5}
    \scalebox{0.90}{
        \begin{tabular}{c|c|c|c|c|c} \hline
            
         \toprule $\alpha_{1}$ & $\alpha_{2}$ & $\alpha_{3}$ & Accuracy & AUC & F1 score \\\hline
          35 & 35 & 30	& 95.09 & 95.37 & 66.88 \\
          35 & 15 & 50	& \textbf{95.71}	& 94.92	& 72.61\\
          25 & 25 & 50  & 95.68	& 95.38	& \textbf{73.52}	\\
          15 & 35 & 50	& 95.58	& 95.32	& 71.15	\\
          20 & 10 & 70  & 95.54	& \textbf{95.79}	& 70.46	\\
          15 & 15 & 70	& 95.48	& 95.54	& 71.46	\\
          10 & 20 & 70  & 95.34	& 95.42	& 69.36	\\
            \hline
        \bottomrule
        \end{tabular}
    }
\end{table}

\subsubsection{Effect of combination of classifiers on pseudo-label prediction}
\label{sec:ensemle_pl_prediction_effect}
Accurate pseudo-label prediction is an essential criterion for the success of our approach. The benefit of having a weighted combination of classifiers for predicting pseudo-labels of reliable unlabeled samples is evident in Table~\myref{tab:ensemble_pl_table}. When all the three classifiers, \textit{i.e.}, similarity classifier, KNN classifier, and linear classifier, are used for estimating pseudo-label, performance (in terms of Accuracy, AUC, and F1 score) improves. Our results indicate that having a weighted combination of classifiers can effectively address the issue of confirmation bias in estimating pseudo-labels, which is present when a linear classifier is used alone. Interestingly, including a linear classifier in the weighted combination of classifiers yields better results than not using it. Fig.~\myref{fig:reliability analysis Figure} compares the percentage of correctly predicted pseudo-labels for a set of reliable unlabeled samples between our approach and the baseline method across different values of $\gamma_1$. The figure illustrates that the performance of the baseline method significantly degrades in predicting pseudo-labels for minority classes due to confirmation bias. In contrast, combining classifiers for pseudo-labeling effectively handles this issue.

\subsubsection{Effect of $\alpha_1$, $\alpha_2$ and $\alpha_3$ in SPLAL pseudo-labeling}
\label{sec:WeightOfClassifierForPL}
Our SPLAL approach uses a weighted combination of a similarity classifier, a KKN classifier and a linear classifier to predict pseudo-labels of reliable unlabeled samples. As described in Sec.~\myref{sec:pseudo_labeling}, the hyperparameters $\alpha_1$, $\alpha_2$, and $\alpha_3$ play a crucial role in the performance of our approach. Table~\myref{table:ablation5} describes the impact of different combinations of $\alpha_1$, $\alpha_2$, and $\alpha_3$ on the performance (in terms of Accuracy, AUC, and F1 score). It is observed that the performance (in terms of AUC) is better when the weight corresponding to the similarity classifier ($\alpha_3$) is high. However, the performance (in terms of F1 score) is relatively better when $\alpha_3$ is relatively low. This finding highlights the importance of label smoothing achieved by incorporating the predictions of the linear classifier and KNN classifier using weights $\alpha_1$ and $\alpha_2$, respectively.
We selected the final values of $\alpha_1$, $\alpha_2$, and $\alpha_3$ as 0.20, 0.10, and 0.70, respectively, giving higher preference to the AUC metric for evaluation. However, these values can be tweaked as per the requirements.

\subsubsection{Comparison with baseline using confusion matrix and ROC curve}
We compare the performance of our approach with the baseline method using confusion matrix and ROC curve. Fig.~\myref{fig:heatmap} compares the confusion matrix obtained by our method and the baseline method over the test dataset. We can see that the baseline method is highly biased and classifies most samples to the majority class. The baseline method cannot correctly predict even a single sample of the disease Benign Keratosis-like Lesions (BKL) and Dermatofibroma (DF). Except for Melanocytic Nevi (NV), correct predictions for all other classes are significantly less. On the other hand, our approach is not biased towards a particular class and performance significantly well for each class.

Fig.~\myref{fig:rocCurveFigure} compares the ROC curve over the test dataset generated by the baseline method and our approach. The performance (in terms of AUC) of the baseline method  for some of the classes, such as Melanoma (MEL), Benign Keratosis-like Lesions (BKL), and Dermatofibroma (DF), is very low as compared to the other classes. In contrast, our approach achieves significant AUC for all the classes.

\subsubsection{Qualitative comparison with baseline using Grad-CAM}
\label{sec:visualization}
We generate visualizations using Grad-CAM to understand the improvement achieved by our method over the baseline method. Fig.~\myref{fig:Gradcam} compares the Grad-CAM \citeblue{jacobgilpytorchcam} visualization between our method and the baseline method. Grad-CAM images are commonly used to locate discriminating regions for object detection and classification tasks. We can see that the baseline method is not able to correctly use the distinguishing features of some of the diseases, such as Melanoma (MEL), Benign Keratosis-like Lesions (BKL), and Basal Cell Carcinoma (BCC), for prediction. However, our method can correctly identify the distinguishing feature of all the seven diseases and use it effectively for classification.

\section{Conclusion}
\label{sec:discussion}
In this work, we propose the Similarity-based Pseudo-Labeling with Alignment Loss (SPLAL) method. SPLAL is a novel SSL approach that aims to improve the classification performance of deep learning models on medical image datasets with limited labeled data availability and class imbalance in its distribution. We propose a novel reliable sample selection method, where we select a set of reliable unlabeled samples, using the similarity with class prototypes criterion. We maintain prototype of every class using the recently viewed training samples. We use a novel method for pseudo-label prediction using a combination of a similarity classifier, a KNN classifier, and a linear classifier. Using a weighted combination of classifiers to estimate high-quality pseudo-labels and incorporating an alignment loss term in the loss function, we aim to improve the model's performance, particularly for minority classes. We extensively evaluate the effectiveness of our approach on two public datasets- the ISIC 2018 and BCCD. Our approach outperforms several state-of-the-art SSL methods across various evaluation metrics, and our ablation studies validate the contribution of different components of our approach.

\section*{Acknowledgments}
We acknowledge the National Supercomputing Mission (NSM) for providing computing resources of \textbf{PARAM Ganga} at the Indian Institute of Technology Roorkee, which is implemented by C-DAC and supported by the Ministry of Electronics and Information Technology (MeitY) and Department of Science and Technology (DST), Government of India.

\section*{Declaration of competing interest}
The authors declare that they have not encountered any financial or interpersonal conflicts that could have an impact on the research presented in this study.


\section*{References}
\bibliography{refs}

\end{document}